\documentclass{article}
\usepackage{titling}
\usepackage{indentfirst}
\usepackage{graphicx}
\usepackage{tikz}
\usepackage{pgfplots}
\usepackage{float}
\usepackage{amsmath}
\usepackage{amssymb}
\usepackage{amsthm}
\usepackage{mathtools}
\usepackage{bm}
\usepackage{amsfonts}
\usepackage{enumerate}
\usepackage{psfrag}
\usepackage{multicol}
\usepackage{color}
\usepackage{url}
\usepackage{algpseudocode}
\usepackage{pythonhighlight}
\usepackage{xy}
\input xy
\xyoption{all}

\newcommand{\RR}{{\mathbb R}}
\DeclareMathOperator*{\argmin}{arg\,min}
\date{}
\title{Compressed Sensing: Mathematical Foundations, Implementation, and Advanced Optimization Techniques}
\author{Shane Stevenson and Maryam Sabagh}
\date{}

\begin{document}

\maketitle

\section{Abstract}
Compressed sensing is a signal processing technique that allows for the reconstruction of a signal from a small set of measurements. The key idea behind compressed sensing is that many real-world signals are inherently sparse, meaning that they can be efficiently represented in a different space with only a few components compared to their original space representation. In this paper we will explore the mathematical formulation behind compressed sensing, its logic and pathologies, and apply compressed sensing to real world signals.
\section{Introduction}
From what humans have studied of the universe so far, we believe that a set of rules govern every interaction in our world. These rules create regularity, and regularity implies sparsity. By leveraging the inherent sparsity of our world’s data, we can derive the surprising ability to
understand patterns with little data.

Previously, the Nyquist limit determined that the sampling rate of a signal must be two times as fast as the fastest frequency within the signal, or else the full signal would not be understood. With recent advances in sparse optimization and computation, a new method has been found that has shattered the nyquist limit. Compressed Sensing leverages the assumed inherent sparsity of a signal to allow for signal reconstruction with far fewer samples than asserted by the Nyquist limit. 

This research paper aims to delve into the core mathematical principles of compressed sensing such as the role of sparsity, optimization, and signal processing. Specifically, we will be exploring how a high dimensional and compressible signal $x\in \RR^n$, can be expressed as a sparse vector $s \in \RR^n$, which means this vector will contain mostly zeros in a transform basis $\Psi \in\RR^{n\times n}$:
\begin{equation}
    x = \Psi s
\end{equation}

The key idea behind compressed sensing is that sparsity allows for more efficient representation of signals while reducing the number of sampling requirements. This is because in a sparse representation, only a small subset of coefficients contributes significantly to the signal's structure.

We will also be exploring practical examples of compressed sensing, and show how to use compressive sensing with modern
technology. Using Python, we will show how easily the Nyquist limit can be beaten, and just
how far we can exploit the natural sparsity of data. Finally, this paper wishes to explore how compressed sensing has evolved with modern computation/optimization techniques, and compare and contrast to more rudimentary methods.

\section{How Does Compressed Sensing Work}
Before jumping straight into the math, it is of utmost importance to fully understand and comprehend why sparsity is so important. While it may not seem obvious at first, the key to compressed sensing is the ability to find a sparse representation of a given vector $x$. This is because if the vector $x$ can be represented by a sparse vector, that means all the data within $x$ can be explained by only a few components in a sparse space. This is exactly how many modern image compression techniques work, like JPG, which represents images as a sparse set of Discrete Cosine Transform coefficients. So, compressed sensing aims to leverage the concept of sparsity to obtain the important components of data within its sparse space, and use those to approximate more data.

Thus, we will express a high dimensional and compressible signal $x\in \RR^n$, as a sparse vector $s \in \RR^n$, in a transform basis $\Psi \in\RR^{n\times n}$:
$$x = \Psi s$$
Taking a small measurement of $x$, denoting it as $y=Cx$, the measurement can  be expressed as:
\begin{equation}
    y=C \Psi s
\end{equation}

\begin{figure}[H]
    \centering
    \includegraphics[scale=.35]{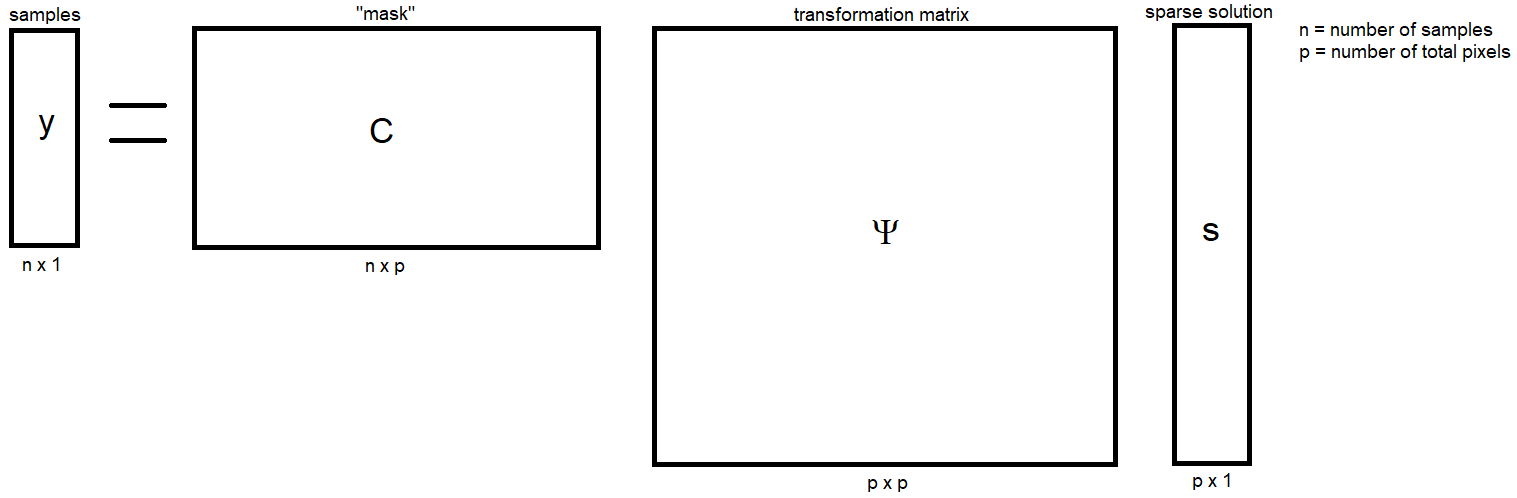}
    \caption{$y=C \Psi s$}
    \label{fig:enter-label}
\end{figure}

\begin{python}
#Read the image
im = io.imread('static/smallDog.png', as_gray=True)

#Flatten the image matrix to an m*n x 1 vector
vecIm = im.flatten()
#Generate random indices
rng = np.random.default_rng()
randIndxs = rng.choice(len(vecIm), int(len(vecIm)*.2), replace=False)
#Take random samples
samples = vecIm[randIndxs]

#Create a mask
mask = np.zeros(vecIm.shape)
mask[randIndxs] = 1

#Multiply the mask with the image element wise to give us y
maskIm = vecIm * mask

#Graph the image and the random measurements we took
fig, axes = plt.subplots(1, 2, figsize=(20, 10))
ax = axes.ravel()

ax[0].imshow(im, cmap=plt.cm.gray)
ax[1].imshow(maskIm.reshape(im.shape), cmap=plt.cm.gray)
\end{python}
\begin{center}
\includegraphics[scale=.4]{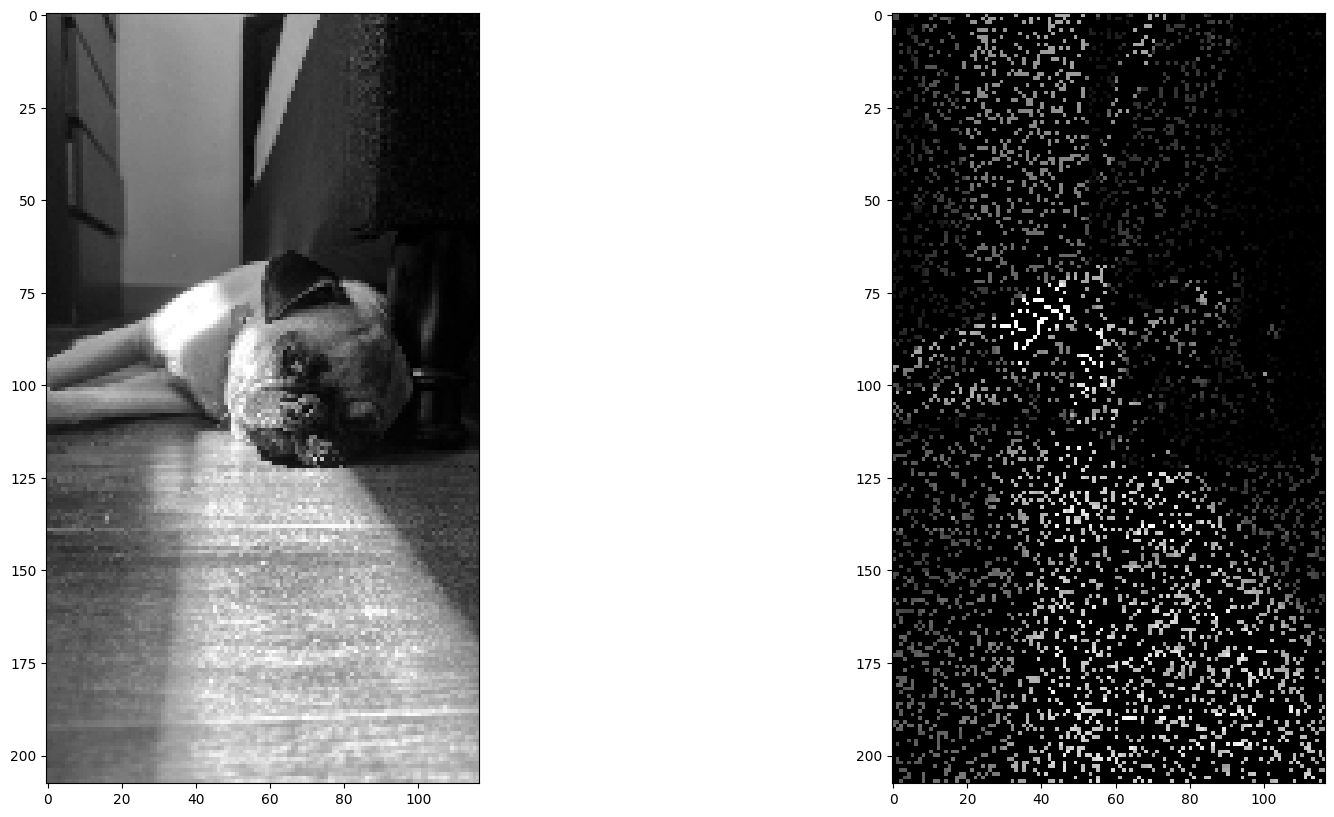}
\end{center}
In this code snippet, we take a natural image, grab 20\% of the pixels at random, and display both the original image, and the data we took from it. This is the data we will first examine compressed sensing with.

Now that we have our $y$ vector, we can begin solving our equation
$$y=C \Psi s$$

We will refer to the matrix $C\Psi$ as $\Theta$.  This system of equations is under determined since $s$ has more entries than $y$, that is, the data we are trying to approximate has more entries than our random samples. Thus, there will be infinitely many $s$ vectors that satisfy the equation $y=\Theta s$. The goal is solve for the sparsest vector $s$ that is consistent with the measurement equation $y=\Theta s$. The key to understanding this equation is to realize that  we are solving for $s$ within a space where we believe the data to be sparse. As long as the real data vector $x$ is sparse within this space, we will be able to find a sparse vector $s$ that can approximate the data $x$ when transformed back into the original space, in this case, image space.

Once we have our random samples, and a space we believe them to be sparse in, we must find a sparse solution out of the infinitely many solutions that exist. This can be characterized by the following equation:
\begin{equation}
    \hat{s}=\argmin_s \|s\|_0 \text{ subject to } y=\Theta s
\end{equation}
where the $\ell_0$ norm represents the number of non-zero elements within a vector. But since using the $\ell_0$ norm will be combinatorially hard, we can optimize with the $\ell_1$ norm instead with the following equation:
\begin{equation}
\hat{s}=\argmin_s \| \Theta s-y\|_2 + \lambda\|s\|_1  
\end{equation}


\subsection{$\ell_p$ Norm Theory}

In compressed sensing, the $\ell_0$ norm and $\ell_1$ norm are crucial concepts that play critical roles in sparsity. The $\ell_0$ norm of a vector is the number of non-zero elements. Using the $\ell_0$ norm, we can find the sparest $s$ vector with the following optimization problem: 
$$\hat{s}=\argmin_s \|s\|_0 \text{ subject to } y=\Theta s$$  The $\ell_0$ norm is not a true norm since it does not satisfy all the norm axioms so this makes using the $\ell_0$ norm combinatorially hard and challenging to work with directly. An alternative to the $\ell_0$ norm is the $\ell_1$ norm. The $\ell_1$ norm is a true norm that is defined as the sum of the absolute value of the vector elements. 
Using the $\ell_1$ norm along with LASSO optimization we can find a sparse $s$ using the following:
\begin{equation}
    \hat{s}=\argmin_s \| \Theta s-y\|_2 + \lambda\|s\|_1 
\end{equation}

 As we know, the concept of sparsity is critical to compressed sensing, therefore critical to understanding and executing compressed sensing is to have a proper understanding of why the $\ell_1$ norm promotes sparsity. Consider the following; take a vector $z\in\RR^n$ where $||z||_2 > b$ with $b > 0$. Now consider the task of making $||z||_2 < b$ by iteratively subtracting or adding to elements of $z$ such that for each iteration, $\sum_{i=0}^{n} |z_i| - |\hat{z}_i| \leq k$ where $\hat{z}$ is the resultant vector after iteration and $k$ is some positive scalar. In text, this means that each iteration we can subtract or add to elements of $z$ such that the sum of the differences between the elements of $z$ and $\hat{z}$ is less than some scalar $k$. In order to achieve $||z||_2 < b$ in the least amount of iterations, each iteration would prioritize shrinking the largest values as, due to the square within the $\ell_2$ norm, shrinking the largest elements of $z$ is more beneficial than shrinking small elements. This results in a final vector $z_*$ where no elements are 0 due to the prioritization of shrinking larger elements and ignoring small ones.
 
 Now consider the same task but with the $\ell_1$ norm. That is, with a vector $z\in\RR^n$ where $||z||_1 > b$ with $b > 0$, make $||z||_1 < b$ by iteratively subtracting or adding to elements of $z$ such that for each iteration, $\sum_{i=0}^{n} |z_i| - |\hat{z}_i| \leq k$. In the $\ell_1$ case, there is no need to prioritize large values becaue there is no square in the $\ell_1$ norm. Thus, the resultant vector $z_*$ can have elements equal to 0 because no elements would be prioritized over others. In other words, adding the $\ell_1$ normalization term onto equation 5, encourages shrinkage of all elements of $s$, which in practice, results in a sparse vector $s$.

 Equation (5) leads to a convex optimization problem, which means finding the sparse $s$ will not be combinatorially hard. LASSO minimizes the sum of squared differences between the sampled data and the transformed $s$ vector, subject to a constraint on the $\ell_1$ norm of $s$.
Using the $\ell_1$ norm will promote sparsity,
but in order to use the $\ell_1$ norm to successfully find a sparse $s$, there needs to be enough measurements and the measurements in $y$ must be incoherent/consist of random measurements. The number of measurements taken, denoted as $p$, will be dependent on the number of non-zero coefficients in our $s$ vector, denoted as $K$. For a $K-$sparse signal in our basis $\Psi$, $p$ can be estimated by:
\begin{equation}
    p \approx k_1K \log(n/K)
\end{equation}
where $k_1$ is a constant multiplier. The constant multiplier, $k_1$, depends on how incoherent our measurements are. In our case, measurements are incoherent when the rows of $C$ have small inner products with the columns of $\Psi$. This is because when the inner product between those elements is small, this means the elements are uncorrelated, meaning the measurements and the basis functions are inherently bringing in different parts of our signal. Having this incoherence in our signal is key for reconstruction. 

\begin{figure}[H]
    \centering
    \begin{tikzpicture}[scale=1]
        \draw[->] (-2.5,0) -- (2.5,0) node[right] {$x_1$};
        \draw[->] (0,-2.5) -- (0,2.5) node[above] {$x_2$};
        \draw[->,red] (0,0) -- (2,0) node[below right] {$\mathbf{v}_1$};
        \draw[->,red] (0,0) -- (0,2) node[above right] {$\mathbf{v}_2$};
        \draw[->,red] (0,0) -- (0,-2);
        \draw[->,red] (0,0) -- (-2,0);
         \filldraw[fill=blue] (0,1) circle (2pt)  ;
        \draw[dashed] (-1,1.5) -- (2,0)  ;

    \end{tikzpicture} 
    \caption{Illustration of the ${\ell_0}$ Norm and its Role in Compressed Sensing}
    \label{fig:l0_unit_ball}
\end{figure}
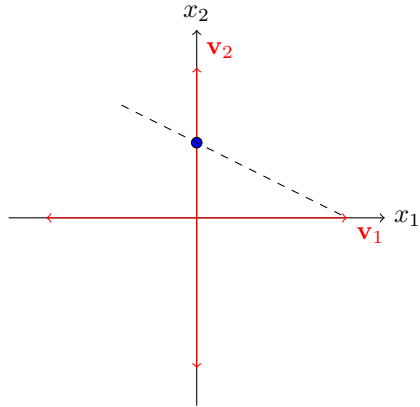

 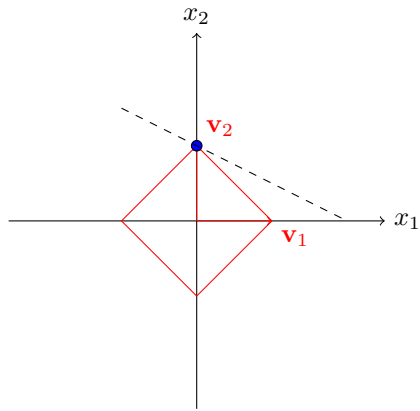
\begin{figure}[H]
    \centering
    \begin{tikzpicture}[scale=1]
        \draw[->] (-2.5,0) -- (2.5,0) node[right] {$x_1$};
        \draw[->] (0,-2.5) -- (0,2.5) node[above] {$x_2$};
        \draw [red](0,1) -- (1,0) -- (0,-1) -- (-1,0) -- (0,1);
        \draw[->,red] (0,0) -- (1,0) node[below right] {$\mathbf{v}_1$};
        \draw[->,red] (0,0) -- (0,1) node[above right] {$\mathbf{v}_2$};
        \filldraw[fill=blue] (0,1) circle (2pt)  ;
        \draw[dashed] (-1,1.5) -- (2,0)  ;
    \end{tikzpicture} 
    \caption{Illustration of the ${\ell_1}$ Norm and its Role in Compressed Sensing}
    \label{fig:l1_unit_ball}
\end{figure}

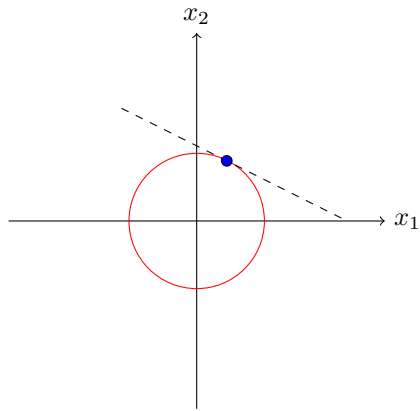
\begin{figure}[H]
   \centering
    \begin{tikzpicture}[scale=1]
        \draw[->] (-2.5,0) -- (2.5,0) node[right] {$x_1$};
        \draw[->] (0,-2.5) -- (0,2.5) node[above] {$x_2$};
        
        \draw[color=red] (0,0) circle (0.9cm);
        \draw[dashed] (-1,1.5) -- (2,0)  ;
        \filldraw[fill=blue] (0.4,0.8) circle (2pt)  ;

    \end{tikzpicture} 
    \caption{Illustration of the ${\ell_2}$ Norm and its Role in Compressed Sensing}
    \label{fig:l2_unit_ball}
\end{figure}

The dashed line represents the
solution set of our under-determined system of equations, equation (2), with different $\ell_p$ balls. We see that the $\ell_0$ norm has the intersection along the axis. This would yield the sparest solution 
since this intersection represents a vector that has the minimum number of non-zero elements.
Similarly, the $\ell_1$ ball has a corner along the axis and the intersection also lies along the axis. Thus, the $\ell_1$ norm minimization will also yield a sparse solution. On the other hand, taking a look at the $\ell_2$ norm minimization, we see a ball shape. The intersection does not lie along the axis so using the $\ell_2$ norm does not yield the sparsest solution.

    

\subsection{Building and Executing Compressed Sensing}
Now that we have a strong understanding of why compressed sensing works, and the role that the $\ell_1$ norm plays in our optimization problem, the next step is to build the optimization problem iteslf. We already have $y$, which is our randomly sampled measurements, but how can we build the $\Theta$ matrix for our optimization equation. Key to doing this, is having a strong understanding of the $\Theta$ matrix, so let's begin understanding it first. If you'll recall, we let $\Theta$ be $C\Psi$, where $\Psi$ 
is the transformation matrix that brings our measurements from a space we believe them to be sparse back to the original domain of the data. According to Brunton and Kutz, $C$ is the measurement matrix that represents a set of $p$ linear measurements on the state of $x$. In other words, $C$ represents the random indices we used to sample $y$ from $x$. 
\begin{figure}[H]
    \centering
    \includegraphics[scale=.4]{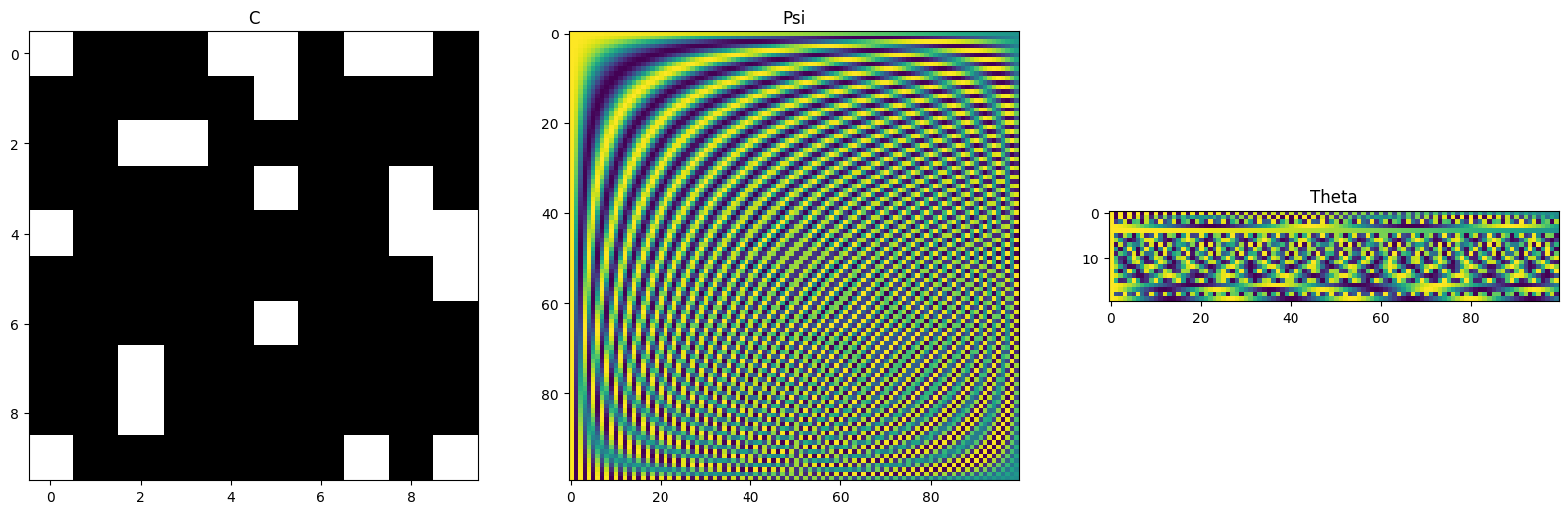}
    \caption{Caption}
    \label{fig:enter-label}
\end{figure}

Thus, $\Theta$ is a matrix that can be multiplied by a solution candidate $s$ to yield a vector that exists in the same space as our sampled data. So, in the context of image reconstruction, a space where natural images are classically sparse in its frequency space. To get to frequency space, we can perform the discrete cosine transform on images where each value of the matrix is a pixel of the image. So, we have a space where our data is sparse, all that's left is to build our $\Theta$ matrix. In the code below, $\Theta$ consists of a set of rows of $\Psi$ which is created by executing the discrete cosine transform on the identity matrix. The specific rows that are selected are based on which random pixels we sampled. For example, say we sampled random pixels 5, 92 and 540, $\Theta$ would consist of rows 5, 92, and 540 of the $\Psi$.

\begin{python}
#Generate an identity matrix where n = # of pixels in the image
id = np.identity(len(vecIm))

#perform the discrete cosine transform
psi = dct(id)

#Sample rows with the same indices used to sample pixels
theta = psi[randIndxs]
\end{python}

\begin{center}
\includegraphics[scale=.8]{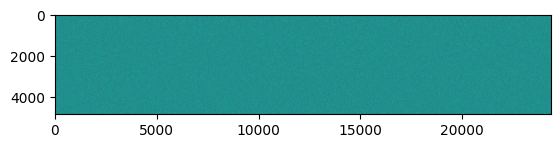}
\end{center}

Once our equation $\hat{s}=\argmin_s \| \Theta s-y\|_2 + \lambda\|s\|_1$ is created, all that's left is to optimize. For this task, LASSO optimization will converge at a sparse solution. Once a sparse $s$ has been converged at, converting $s$ back into the original data domain will hopefully result in a vector that accurately represents the original data $x$. In our case, this means performing the inverse discrete cosine transform on our $s$ vector to create an approximation of our original image

\begin{python}
#Perform Lasso regression in order to optimize for a sparse s
res = Lasso(alpha=.0001).fit(theta, samples.reshape(-1, 1))

#Grab the components of s
s = res.coef_

#Perform the inverse discrete cosine transform to return from the spectral domain to image space
r = idct(s)

#Display the reconstructed image
plt.imshow(r.reshape(im.shape), cmap = plt.cm.gray)
\end{python}
\begin{figure}[H]
    \centering
    \includegraphics[scale=.8]{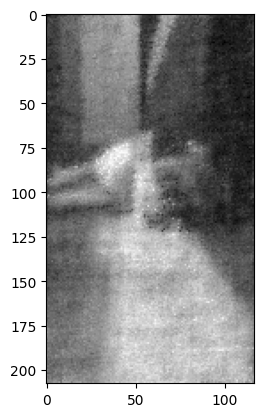}
    \caption{Reconstructed Image}
    \label{fig:enter-label}
\end{figure}

And there you have it! While the reconstructed image is far from perfect, it is still quite surprising how well an image can be reconstructed from only a portion of the original pixels. All that said, this is not very satisfying. The method described so far scales very poorly for larger images, and in order to perform compressed sensing on this image, it was actually required to downsample the image otherwise it would require too much memory, and would be far too slow. Specifically, performing the discrete cosine transform on such a large matrix is very cumbersome. In the following section, we will examine an alternative to Lasso regression in order to make compressed sensing much more scalable, and satisfying.

\section{Advanced implementation }

As mentioned before, our python implementation, while functional, is far from efficient. Firstly, performing the discrete cosine transform on a matrix with as many entries as there are pixels in the image squared is simply too large to live in RAM. By leveraging very efficient C code we can store this matrix deeper in our computer's memory, and simply pull out the correct entries when needed. Furthermore, because Lasso leverages coordinate descent, a history of previous iterations must be kept in memory in order to quickly converge at a solution. The solution to this: Orthant-Wise Limited-Memory Quasi-Newton. The aforementioned mouthful, abbreviated simply as OWL-QN, is built off of centuries of theory, and cannot be introduced without thorough explanation. So, starting from the foundations of Newton's Method, we will build our understanding of Quasi-Newton algorithms in order to understand what makes this optimization routine so much better than our naive approach.

\subsection{Newton's Method}
First introduced in the 18th century, Newton's method is a fairly naive root finding algorithm. The algorithm works by iteratively producing better approximations of the roots of a function. The most basic version can be described by the following equation:
\begin{equation}
    x_{n+1} = x_n - \frac{f(x_n)}{f'(x_n)}
\end{equation}

Where $f$ is a real valued function. This can be taken one step further, and we can consider using newton's method on the derivative of $f$ in order to find the extrema of $f$. In other words, our root finding algorithm can become an optimization routine, and can help us locate the minimum of $f$ with the following:

\begin{equation}
    x_{n+1} = x_n - \frac{f'(x_n)}{f''(x_n)}
\end{equation}

\subsection{Higher Dimensions and Quasi-Newtons}
Now consider how to apply Newton's Method to a vector valued function. The higher dimensional extension comes relatively simply. In this case, the derivative of $f$, $f'$, becomes the gradient of $f$, $\nabla f$, and the second order derivative of $f$, $f''$, becomes the Hessian of $f$, $H$.

\begin{equation}
    x_{n+1} = x_n - H(x_n)^{-1}\nabla f(x_n)
\end{equation}

This successive updating algorithm, given a few conditions like a good enough initial guess, will converge at a minimum of $f$. However, when put to computation, large vectors imply large hessian matrices, which implies poorly costly calculations. In other words, the Hessian will be of size $n x n$, so calculating $n^2$ second derivatives is not viable for large $n$'s This is when we enter the realm of Quasi-Newton algorithms. Quasi-Newton algorithms leverage the fact that the if the gradient is already computed, the hessian can be approximated by a linear number of scalar operations, allowing Quasi-Newton algorithms to execute quickly even for large $n$.

\subsection{BFGS}
Quasi-Newton algorithms are characterized by the different ways through which $H$, and conversly $H^{-1}$ are calculated. The specific Quasi-Newton function we will be looking at is the Broyden–Fletcher–Goldfarb–Shanno (BFGS) algorithm. The BFGS algorithm computes successive Hessians with the following algorithm:

\begin{equation}
    H_{k+1} = \frac{y_ky_k^T}{y^T\Delta x_k} - \frac{H_k \Delta x \Delta x^T H_k}{\Delta x^T H_k \Delta x}
\end{equation}

Where $y_k$ represents the gradient at the previous iteration's function value, and $\Delta x$ represents the difference between the current $\Vec{x}$ and the previous $\Vec{x}$. In other words, $\Delta x_k = \Vec{x}_{k+1} - \Vec{x}_k$. This algorithm is organized into a recursive function who's inputs are the past histories of all gradients ${y_k}$, and all the prior function inputs ${x_k}$. Once the hessian is approximated, the inverse hessian, $H_k^{-1}$ is approximated by applying the Woodbury formula, or the Sherman-Morrison formula, a special case of the Woodbury formula. Ultimately, these formulas and their proofs are outside the scope of this text, but are well documented online.

Because we must record the history of our inputs and gradients, and compute hessian approximations based on them, both this function's runtime and memory requirement will scale $O(n^2)$. Improving upon the method's runtime proves to be quite difficult, but improving on its memory requirement is not quite as challenging.



\subsection{L-BFGS}
Although the addition of a hessian approximation allows for the runtime of quasi-newton functions to scale $O(n^2)$, the memory requirement, also $O(n^2)$ causes a big problem for most computation. This is because the entire history of function inputs and gradients is still being stored in memory, and for large n's, even quasi-newtons will fail. Enter the Limited memory BFGS, or L-BFGS for short. In order to cut back on the memory needed for the full BFGS Quasi-Newton, the L-BFGS only stores the last m inputs and gradients, where m is a value chosen by the user. This causes the hessian approximation to be worse, but allows the memory to scale $O(m \cdot n)$

\subsection{Orthant-Wise Limited-Memory Quasi-Newton}
Up until this point we have not looked at how quasi newtons interact with the $\ell_1$ norm. Thats because they don't. The $\ell_1$ norm causes issues with the functions differentiability, and causes even the L-BFGS to fail. The OWL-QN algorithm is a variant of the L-BFGS algorithm that optimizes a function of the form $f(\Vec{x}) = g(\Vec{x}) + ||\Vec{x}||_1$ where $g$ is a differentiable, convex loss function. At each iteration, OWL-QN estimates the sign of each component in $\Vec{x}$ , and restricts the subsequent step to have the same sign. This causes the $||\Vec{x}||_1$ term to become a smooth linear term, one that the L-BFGS algorithm can handle. Furthermore, after an L-BFGS step, the signs of the vector $\Vec{x}$ can be re-estimated.

\subsection{results}
Using a C implementation of the OWL-QN algorithm allows for a much larger image to be processed. The code to achieve this will be listed on the github repository linked in the reference section. Furthermore, OWL-QN saves enough memory to allow us to perform computation with all three image channels, so we can reconstruct the image with color.

\begin{figure}[H]
    \centering
    \includegraphics[scale=.42]{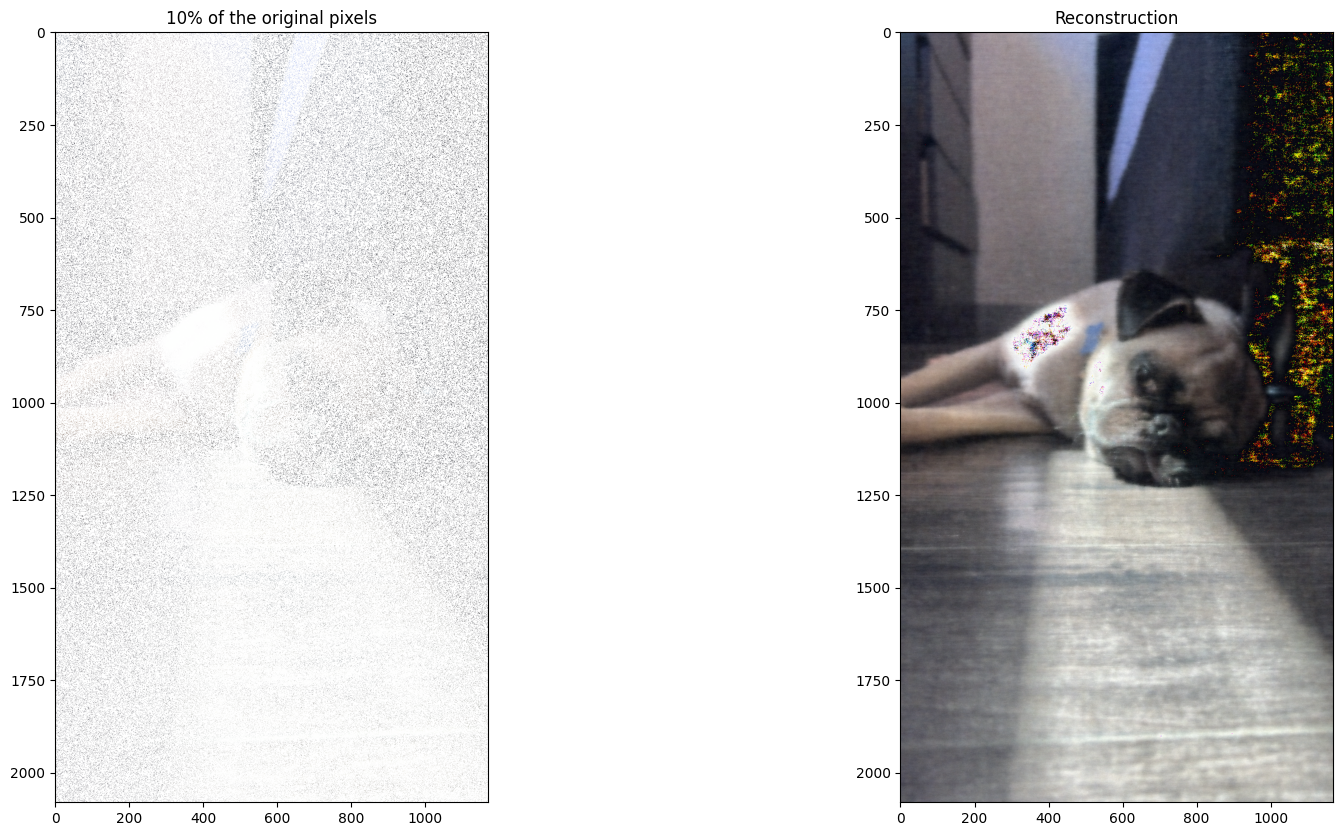}
    \caption{Constructed Image Using \text{10\%} of Original Pixels}
    \label{fig:enter-label}
\end{figure}

\begin{figure}[H]
\centering
    \includegraphics[scale=.42]{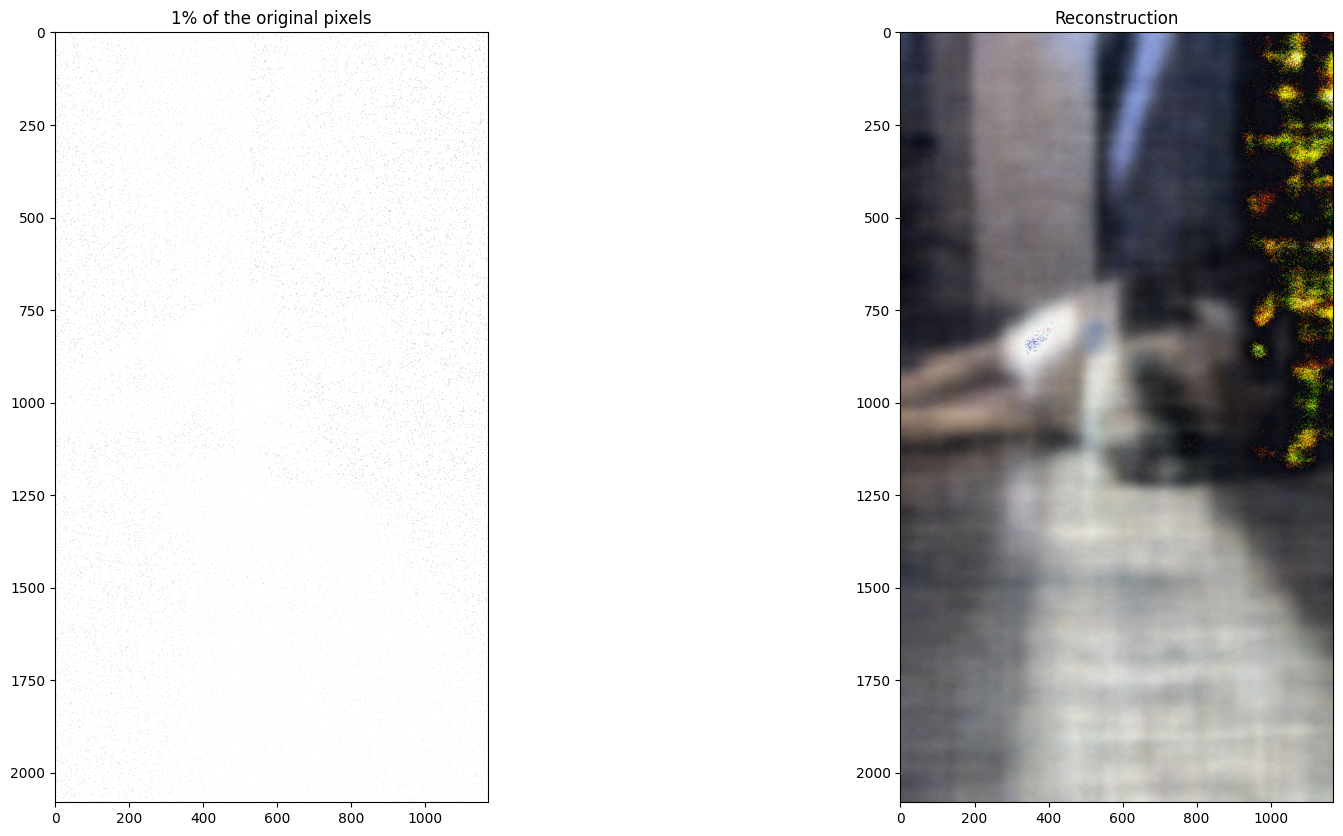}
    \caption{Constructed Image Using \text{1\%} of Original Pixels}
    \label{fig:enter-label}
\end{figure}

While there are a few noticeable errors, mainly held on the right side of the image where the pixels tend to be very close to black, the reconstruction of the full image is quite impressive. The 10\% reconstruction is extremely faithful to the original image, but what might be  most surprising is the 1\% reconstruction. Even with only 1\% of the original data, we can leverage the inherent sparsity in our universe and reconstruct a signal from close to nothing.


\end{document}